\documentclass{article}
\usepackage{graphicx}
\graphicspath{ {./images/} }

\usepackage{float}
\usepackage{arxiv}

\usepackage{appendix}
\usepackage{etoolbox}
\BeforeBeginEnvironment{appendices}{\clearpage}

\usepackage{mathtools,xparse}

\usepackage{tikz}
\usetikzlibrary{arrows,backgrounds}
\usepgflibrary{shapes.multipart}
\usepackage[utf8]{inputenc} 
\usepackage[T1]{fontenc}    
\usepackage{hyperref}       
\usepackage{url}            
\usepackage{booktabs}       
\usepackage{amsfonts}       
\usepackage{nicefrac}       
\usepackage{microtype}      
\usepackage{lipsum}
\usepackage{wrapfig}
\usepackage{sidecap}
\sidecaptionvpos{figure}{c}
\usepackage{xcolor}
\usepackage[linesnumbered,ruled,vlined]{algorithm2e}
\SetKwInput{KwInput}{Input}                
\title{Embedded methods for feature selection in neural networks}

\author{
  Vinay Varma K. \\
  Machine Learning Engineer\\
  Niveshi\\
  New Delhi, PIN 110034 \\
  \texttt{vinay@niveshi.com} \\
  
}

\begin{document}
\maketitle

\begin{abstract}
The representational capacity of modern neural network architectures has made them a default choice in various applications with high dimensional feature sets. But these high dimensional and potentially noisy features combined with the black box models like neural networks negatively affect the interpretability, generalizability, and the training time of these models. Here, I propose two integrated approaches for feature selection that can be incorporated directly into the parameter learning. One of them involves adding a drop-in layer and performing sequential weight pruning. The other is a sensitivity-based approach. I benchmarked both the methods against Permutation Feature Importance (PFI) - a general-purpose feature ranking method and a random baseline. The suggested approaches turn out to be viable methods for feature selection, consistently outperform the baselines on the tested datasets -  MNIST, ISOLET, and HAR. We can add them to any existing model with only a few lines of code.
\end{abstract}


\section{Introduction}
\subsection{Motivation}
Outside of domains like Computer Vision and Natural Language Processing (NLP), identifying relevant input feature set for the task is not obvious. For example, consider the problem of forecasting stock-returns. Here, potentially relevant features might range from the returns of the index and other stocks to the tweets of the CEO. Even in applications where Domain-expertise renders a good feature set, the predictions from the model are rarely the end-goal\footnote{Complete end-to-end decision-making systems aren’t yet possible with machine learning, at least in high stakes environments. For instance, consider using machine learning for decision making in healthcare\cite{rl}. Though, In theory, diagnosing and treating a patient can be formulated under active reinforcement learning, present machine learning models only augment the expertise of the human doctor.}. The work aims to present two simple methods that reliably give feature importance ranking in neural networks. It’s important to note that the method needs to be embedded - use the specific predictor in this case neural networks and be integrated into the parameter learning. This ensures appropriate interaction effects relevant to the optimization criteria are taken into account and the redundancies thereby are removed. Though the results are presented on feed-forward neural networks in a classification setting, by construction the same methods can be used with any learning criteria and architecture.

\subsection{Background}

For a given dataset $X^{n \times d}$, any Feature ranking method involves returning a list  I = [$f_1$,$f_2$,\dots $f_d$]. The magnitude of the elements in the list is used to rank order the features. Here I detail the Permutation Feature Importance(PFI)\cite{Brieman2001} algorithm which I here  used as a baseline. PFI is a widely used importance attribution technique for random forests. Here, I adapted the model agnostic version\cite{Rudin2019} of the method for neural networks.

\begin{algorithm}[H]
\SetAlgoLined
\DontPrintSemicolon
\KwInput{training data $X  \in  \mathbb{R}^{n \times d}$, training labels $Y$, neural network $f_{\theta}(.)$, number of random permutations $c$, Accuracy measure of the model $A_{f_{w}}(X,Y)$ } 
\For{dim in \normalfont 1,\ldots, d}    
        { 
        Let the feature importance of the $dim$ be $f_c$ \;
        $ f_c \leftarrow 0$ \;
        \For{$i$ in \normalfont 1,\ldots, c }{
        	Generate a new input data $X_{perm}$ with random permutation of the column $dim$ in $X$ \;
        	 $ f_c  \leftarrow  f_c + A_{f_{w}}(X,Y) - A_{f_{w}}(X_{perm},Y)$
        	}
        	The quantity $\frac{f_c}{c}$ gives the feature importance for each $dim$ in $d$ \;
        
        }
\caption{Permutation Feature Importance}
\label{PFI}
\end{algorithm}

PFI has numerous shortcomings, especially under correlated-features \cite{Molner}. Nevertheless, the generality and the simplicity of the method makes it a good baseline.

The contributions of the present work are two-fold:
\begin{enumerate}
  \item A drop-in layer with iterative pruning of the weights is shown to be a viable feature selection method.
      \item Many prediction-explanation \cite{Arras2017,Sudararajan,Shirkumar,Simyon2014} methods are used in neural networks. These are post-hoc instance-wise explainers. This work uses a gradient-based sensitivity method for global feature selection. I also propose that the testing framework used herein, where the features are removed and retrained should be adopted to approximate the relative validity of any feature attribution technique. As the approach doesn't distinguish between instance-specific and global attribution methods, the results would allow us to compare between them and also use them for the task of feature selection.
\end{enumerate}

\section{Proposed Methods}
\label{sec:methods}
\subsection{Sensitivity based Selection (SBS)}
\label{subsec:sbs}
Specific to neural networks several feature attribution methods exist. These approaches typically involve backpropagating an importance signal from the output neuron to the input features. For instance, Simonyan et al. used saliency maps \cite{Simyon2014} for images that involve computing the gradient of the output w.r.t. pixels of an input image. To the best of my knowledge, they are designed as a Visualization-aid and are not used for feature selection. Sensitivity based selection method can be summarized as follows. Considering the classification problem with  input $X$  $\in$  $\mathbb{R}^{n \times d}$. First, train the model on this data. Now, for each instance of the data point $x$ $\in$ $X$, compute the gradient of the output label w.r.t $x$ on this model. Take the absolute value of the average of this gradient across all the data instances. The resulting list of values I = [$f_1$,$f_2$,\dots $f_d$] gives the global feature sensitivity of the model under consideration. Similarly, In a regression setting gradient can be taken w.r.t the error signal. The intuition is that the absolute value of the partial derivative of input feature w.r.t relevant output neuron, In a  classification setting its the corresponding class-probability, quantify its importance. This method can be seen as adapting PFI for the special case of neural networks, using the gradient information. PFI, being model agnostic randomly permutes a feature and uses the drop in accuracy (or another similar metric) as a proxy for feature importance. Moreover, with PFI computational cost scales linearly with the dimensionality of the dataset. Neural Networks implemented with libraries like Pytorch\cite{Pytorch} allows the computation of this feature sensitivity in a single forward and backward pass through the network.

\subsection{Stepwise Weight Pruning Algorithm (SWPA)}
\label{subsec:swpa}
In Neural networks, weight-pruning refers to systematically removing the parameters of the existing network. For a given reduction in parameters, pruning techniques aim to minimize the loss in the performance of the original model.\footnote{Performance doesn't necessarily always decrease with pruning. see \cite{LTH}}  Here I use a drop-in layer that can be incorporated into any neural network and sequentially prune its weights to arrive at the most important ones. The idea is if these weights $W$ $\in$ $\mathbb{R}^{1 \times d}$ are in the first layer of the network and are all initialized to ones and the output of the first layer $O = \{w_1.x_1,\dots ,w_d.x_d\}$ is the multiplication of the corresponding elements of input, then pruning a weight $w_i$ has a direct interpretation of removing $x_i$. The complete algorithm is summarized below.

\begin{algorithm}[H]
\label{alg:alg2}
\SetAlgoLined
\DontPrintSemicolon
\KwInput{training data $X$  $\in$  $\mathbb{R}^{n \times d}$, 
training labels $Y$, base network $f_{\theta}(.)$, Drop-in Layer $W$ , Step Counter $n$ $\in$ $\mathbb{Z}_{\geq 1}$ , Selection factor $f \in [0,1]$}
\For{count in \normalfont 1,\ldots, $n+1$}    
        { 
        $O \leftarrow \{w_1.x_1,\dots ,w_d.x_d\}$\;
        \If{$count > 1$}{
  $k \leftarrow \frac{(1-f)*d}{n}$\;
  Sort the weights $W$ of the Drop-in Layer based on their absolute value.\;
  Set the least $k$ of them to 0. \;
  
   }
   Train the base network on $O$ \;
        }
    Take the features corresponding to the top $f$ fraction of the weights in $W$ based on their absolute value and train them on the base network.

\caption{Stepwise Weight Pruning Algorithm (SWPA)}
\end{algorithm}

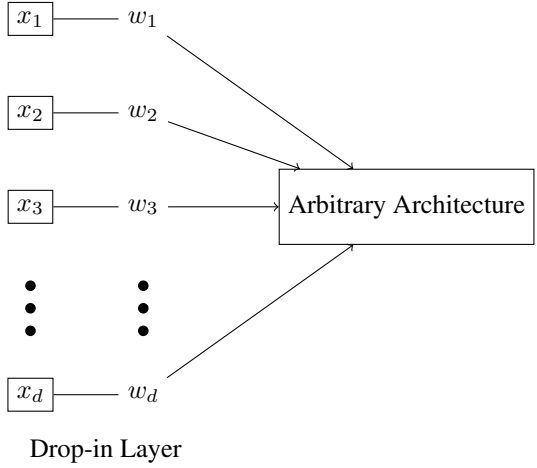
\begin{SCfigure}

    \begin{tikzpicture}
    	\tikzstyle{place}=[circle, draw=black, minimum size = 8mm]
    	\tikzstyle{rectangle_style}=[rectangle, draw]
        \tikzstyle{arbitrary_arch}=[rectangle , draw, minimum width=0.6cm,
                        minimum height = 1cm]

        \foreach \x in {1,...,3}
            \draw node at (0, -\x*1.25) [rectangle_style] (input_\x) {$x_\x$};
        \foreach \x in {1,...,3}
            \fill (0, -4.5 -\x*0.3) circle (2pt);
        \draw node at (0, -5*1.25) [rectangle_style] (input_d) {$x_d$};
    	
        \foreach \x in {1,...,3}
            \draw node at (1.5, -\x*1.25) [] (w_\x) {$w_{\x}$};
        \foreach \x in {1,...,3}
            \fill (1.5, -4.5 -\x*0.3) circle (2pt);
        \draw node at (1.5, -5*1.25) [] (w_d) {$w_{d}$};
         
        \foreach \x in {1}
            \draw node at (5, -3*1.25) [arbitrary_arch] (rect_\x){Arbitrary Architecture};
             
        
        \foreach \i in {1,...,3}
            \draw [-] (input_\i) to (w_\i);
        \draw [-] (input_d) to (w_d);

        
        \foreach \i in {1,...,3}
            \draw [->] (w_\i) to (rect_1);
        \draw [->] (w_d) to (rect_1);
       
    	
    	\node at (1, -7) [black, ] {Drop-in Layer};
    \end{tikzpicture}
    \caption{\textbf{Modified Network}\\
    The Inputs of the orginal model are first passed through the Drop-in Layer. Then,the weights $w_i$ are dropped sequentially during the training according to \textbf{Alogrithm 2.}}
    \label{fig:Modified Network}

\end{SCfigure}

\section{Experiments}
\label{sec:experiments}
\subsection{Data and Setup}
\label{subsec:setup}

Here, I give a brief description of the datasets used.

\textbf{Smartphone  Dataset for  Human  Activity  Recognition  (HAR)}is  the data collected from smartphone sensors mounted on to the humans performing various activities like WALKING, LYING, STANDING, etc.  For each data instance, 561 features are given.

\textbf{ISOLET} is a widely used speech dataset collected from 150 individuals speaking each letter of the alphabet. The data is preprocessed to include 617 attributes like spectral coefficients; contour and sonorant features.

\textbf{MNIST} is one of the most commonly used datasets in the machine learning community. It consists of 28-by-28 gray-scale images of handwritten digits from 0-to-9. All the images in the dataset are centered. Thus, each pixel can be safely treated as a separate feature.

In all the experiments data is randomly divided into 60-20-20 split between train, validation, and test set respectively. Training is done with a budget of 20000 epochs, and the patience factor is set to 2000 epochs. This essentially means the model is allowed to be trained for up to 20000 epochs only if there is no degradation of the performance on the validation set on any continuos set of 2000 epochs. The best performing model on the validation set is chosen and tested on the test set. A 3 layer feedforward neural network with a reduction factor of 2 is used in all experiments. Complete details of the network architecture are given in Appendix B. SGD optimizer is used with a learning rate of 0.001 and a momentum of 0.9. Other hyperparameters include the Step Counter $n$ of the SWPA which is set to 4, Selection Factor $f$ to 0.1, and the number of random permutations $c$ in PFA to 10.

\subsection{Results}
\label{subsec:results}

In Table~\ref{tab:table} both the top and the bottom 10\% of the features are selected based on the proposed methods and the corresponding accuracies on the test set are presented. Both the methods are able to effectively rank order the input features. Of the two, SWPA outperforms on all the data sets. Table~\ref{tab:table2} summarizes the results on the baseline - Permutation Feature Importance (PFI). As expected, PFI reduces the feature importance when a correlated feature is added \cite{Molner}, and shares it between the two. This explains the small the accuracy gap between the top and the bottom as informative features are spread between both the groups\footnote{ All the used datasets have high degree of correlated features.}. But, It performs better than random assignment, acting as a reasonable baseline. For the accuracies on the randomly chosen features, 10 runs are used.
\begin{table}[H]
 
  \centering
  \begin{tabular}{lllllll}
                   
    \cmidrule(r){1-7}
    Dataset     & (n,d)     & \#feat  & Top SWPA &Btm SPWA  &Top SBS &Btm SBS \\
    \midrule
    HAR &(5744,561)  &56  &\textbf{0.976}  &0.654 &0.932 &0.574     \\
    
    ISOLET &(7797,617) &61  &\textbf{0.900}  &0.411 &0.841 &0.498     \\
    
    MNIST  &(60000,784) &78 &\textbf{0.941} &0.134 &0.880 &0.153\\
    
    \bottomrule \\
  \end{tabular}
  \caption{Classification accuracies on the features chosen by the proposed methods}
  \label{tab:table}
\end{table}

\begin{table}[H]
 
  \centering
  \begin{tabular}{lllll}
                   
    \cmidrule(r){1-5}
    Dataset      & Top PFI &Btm PFI  &Random Worst &Random Avg \\
    \midrule
    HAR   &\textbf{0.917} &0.814  &0.306 &0.504     \\
    
    ISOLET  &\textbf{0.787} &0.700  &0.717 &0.766  \\
    
    MNIST   &\textbf{0.893} &0.867 &0.323 &0.714\\
    
    \bottomrule \\
  \end{tabular}
  \caption{Classification accuracies on PFI based features and the randomly assigned ones.}
  \label{tab:table2}
\end{table}

\section{Ablations}

\subsection{Effect of Step Counter}
\begin{SCfigure}[]
    \includegraphics[scale=0.6]{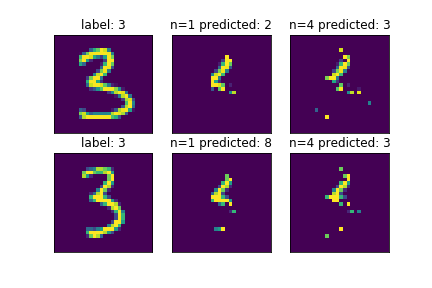}
    \caption{\textbf{Effects of Step Counter $n$}\\
    The second and the third column highlights the final selected features by SWPA ($n=1$) and SWPA ($n=4$) respectively. In both the instances, SWPA ($n=1$) misclassifies the input. Visually, It can be inferred that SWPA ($n=4$) chooses slightly more informative features.}.

    \label{fig:Stepwise}

\end{SCfigure}

In this section, I answer the question Is the Step Counter $n > 1$ (SWPA) adding any value? 
I compared the accuracy of the resulting models with $n=1$ and $n=4$. It is interesting to note that Sequential pruning is adding a noticeable improvement in all the cases. In figure ~\ref{fig:Stepwise} , I have chosen two instances of the MNIST dataset where SWPA (n=1) misclassified them. About 35\% of the features in $SWPA$ ($n=4$) are different from  $SWPA$($n=1$). Given, the performance of the model on $SWPA$($n=4$), It indicates that these features are responsible for the marginal gains.

\begin{table}[H]
 
  \centering
  \begin{tabular}{lll}
                   
    \cmidrule(r){1-3}
    Dataset      &SWPA with $n=4$  &SWPA with $n=1$  \\
    \midrule
    HAR   &\textbf{0.976} &0.949       \\
    
    ISOLET  &\textbf{0.900} &0.835    \\
    
    MNIST   &\textbf{0.941} &0.877 \\
    
    \bottomrule \\
  \end{tabular}
  \caption{SWPA with stepsize = 1}
  \label{tab:table3}
\end{table}

\subsection{Effects of Adding constraints to the objective function }

Here, I tested the effect of adding various constraints to the original objective function $L_{(f_{\theta},W)}$. First I added \textit{l1} regularization on the weights $W$ of the Drop-in Layer. 

\begin{equation}
\label{eqn:l1}
\textit{l1}   = \lambda.\lVert W \rVert_{1}
\end{equation}

\begin{equation}
\label{eqn:wvl}
wvl  = -  \gamma.\mathrm{Var}[Sigmoid(20*W-0.5)]
\end{equation}

Further, to make the weights more spread out and hence distinctive, I also considered adding weight variance loss ($wvl$) as in (2). In all the experiments I chose $n=1$ for simplicity. $\lambda$, $\gamma$ are set to $1$ and $10$ respectively.
\begin{table}[H]
 
  \centering
  \begin{tabular}{lllll}
                   
    \cmidrule(r){1-5}
    Dataset     &base &only $l1$ &only $wvf$  &$l1$ and $wvf$   \\
    \midrule
    HAR   &0.949 &0.963  &0.961   &0.960  \\
    
    ISOLET  &0.835 &0.833  &0.816  &0.828\\
    
    MNIST   &0.877 &0.879 &0.877 &0.867\\
    
    \bottomrule \\
  \end{tabular}
  \caption{Effects of adding constraints - $l1$ and Weight variance factor ($wvf$). Comparision of the features selected in each of these methods is given in Appendix A.}
  \label{tab:table4}
\end{table}

As there is no considerable difference in performance, I conclude that the marginal benefit of adding constraints like $l1$ or $wvf$ is negligible for the task of selecting informative features.

\section{Conclusion}
\label{subsec:conclusion}
As neural network based methods are becoming the default choice for modeling problems in high stakes applications like Finance and Medicine, maintaining the model and being able to reason about its predictions is crucial. Moreover, In real-world applications performance on the validation set or even the test set doesn't give a complete picture as the statistical correlations can always break and new ones can emerge in the future. In this regard, a simpler model with fewer features is always desirable. 
In this work, I proposed two approaches for feature selection in neural networks, of which a simple pruning based method (SWPA) outperformed on all the datasets.

\bibliographystyle{unsrt}  


\begin{appendices}
    
 \section{Feature Similarity on SWPA with various constraints}   
 
 The numbers in the below tables indicate fraction of Features that are similar between the methods. Again, for all comparision the top 10\% of the features are taken.
 \label{sec:AppendixA}
 
  \begin{table}[H]
 
        \centering
        \begin{tabular}{lllll}
                       
        \cmidrule(r){1-5}
        Experiment      &base &$l1$  &$wvl$  &$l1$ and $wvl$  \\
        \midrule
        base   &1.0 &0.803  &0.839  &0.839   \\
        
        $l1$  &0.803 &1.0  &0.857 &0.839  \\
        
        $wvl$ &0.839  &0.857 &1.0 &0.785\\

    \bottomrule \\
  \end{tabular}
  \caption{HAR}
  \label{tab:tableD}
\end{table}

  \begin{table}[H]
 
        \centering
        \begin{tabular}{lllll}
                       
        \cmidrule(r){1-5}
        Experiment      &base &$l1$  &$wvl$  &$l1$ and $wvl$  \\
        \midrule
        base   &1.0 &0.786  &0.852  &0.786   \\
        
        $l1$  &0.786 &1.0  &0.786 &0.852  \\
        
        $wvl$ &0.852  &0.786 &1.0 &0.786\\

    \bottomrule \\
  \end{tabular}
  \caption{ISOLET}
  \label{tab:tableE}
\end{table}

  \begin{table}[H]
 
        \centering
        \begin{tabular}{lllll}
                       
        \cmidrule(r){1-5}
        Experiment      &base &$l1$  &$wvl$  &$l1$ and $wvl$  \\
        \midrule
        base   &1.0 &0.820  &0.807  &0.897   \\
        
        $l1$  &0.820 &1.0  &0.833 &0.807  \\
        
        $wvl$ &0.807  &0.833 &1.0 &0.820\\

    \bottomrule \\
  \end{tabular}
  \caption{MNIST}
  \label{tab:tableF}
\end{table}

\section{Network}
\label{sec:AppendixB}

A 3 layer feed-forward neural network is used with a reduction factor of 2,meaning at each successive layer the dimensionality of the input is reduced by half. The final layer has the dimensionality equal to the number of distinct classes in the dataset. Tabel ~\ref{tab:tableG} and ~\ref{tab:tableH} summarizes the number of neurons in each layer for the networks on both reduced and full feature sets.

  \begin{table}[H]
 
        \centering
        \begin{tabular}{lllll}
                       
        \cmidrule(r){1-5}
        Dataset      &\#feat &\#layer 1  &\#layer 2  &\#layer 3   \\
        \midrule
        HAR           &56     &28      &14    &6   \\
        
        ISOLET  &61 &30  &15 &26  \\
        
        MNIST &78  &39 &19 &10\\

    \bottomrule \\
  \end{tabular}
  \caption{Network on reduced feature sets.}
  \label{tab:tableG}
\end{table}

  \begin{table}[H]
 
        \centering
        \begin{tabular}{lllll}
                       
        \cmidrule(r){1-5}
        Dataset      &\#feat &\#layer 1  &\#layer 2  &\#layer 3   \\
        \midrule
        HAR           &561     &280      &140    &6   \\
        
        ISOLET  &617 &308  &154 &26  \\
        
        MNIST &784  &392 &196 &10\\

    \bottomrule \\
  \end{tabular}
  \caption{Network on full data set.}
  \label{tab:tableH}
\end{table}

\end{appendices}

\end{document}